\pdfoutput=1
\documentclass[conference]{IEEEtran}
\IEEEoverridecommandlockouts
\usepackage{cite}
\usepackage{stfloats}

\usepackage{amsmath,amssymb,amsfonts}
\usepackage{algorithmic}
\usepackage{graphicx}
\usepackage{textcomp}
\usepackage{xcolor}
\def\BibTeX{{\rm B\kern-.05em{\sc i\kern-.025em b}\kern-.08em
    T\kern-.1667em\lower.7ex\hbox{E}\kern-.125emX}}
\begin{document}
\title{Bi-Level Routing and Sparse Spatial Attention based Multi-View BEV 3D Object Detection for Autonomous Driving}

\author{\IEEEauthorblockN{1\textsuperscript{st} Jing Zhang\textsuperscript{*}}
	\IEEEauthorblockA{\textit{Stuart Weitzman School of Design} \\
		\textit{University of Pennsylvania}\\
		Philadelphia, USA \\
		jing321@alumni.upenn.edu\\
		\textsuperscript{*}Corresponding author}
\and
\IEEEauthorblockN{2\textsuperscript{nd}Jiaqi Liu}
\IEEEauthorblockA{\textit{Courant Institute of Mathematical Science} \\
	\textit{New York University}\\
	New York, USA \\
	jl8456@nyu.edu}
	\and
	\IEEEauthorblockN{3\textsuperscript{rd}Zibo Wang}
	\IEEEauthorblockA{\textit{Courant Institute of Mathematical Science} \\
		\textit{New York University}\\
		New York, USA \\\
		zibo.wang@nyu.edu}
}

\maketitle

\begin{abstract}
Bird’s Eye View (BEV)-based multi-view 3D object detection suffer from challenges of computational complexity ,  multi-scale feature extraction, and efficiency of dense 2D-to-BEV view transformation. 
To address these problems, this paper proposes an improved BEV 3D object detection algorithm Sparse-BEVNet. Firstly, a Bi-Level Routing Attention (BRA) mechanism is introduced into the image feature extraction network to reduce the computational burden of the backbone. Second, Cascaded Group Attention (CGA) is employed in the feature fusion module, which enhances deep interaction across features of different hierarchical levels without introducing additional computational overhead.
Furthermore, a Sparse Spatial Cross-Attention mechanism is adopted to replace the conventional dense view projection pipeline. 
Experimental results on the public nuScenes dataset demonstrate that the proposed method achieves a mean Average Precision (mAP) of 45.2\% and a nuScenes Detection Score (NDS) of 54.5\%, corresponding to 3.6\% and 2.8\% improvements relative to the baseline model, respectively.
\end{abstract}

\begin{IEEEkeywords}
3D Object Detection; Bird's Eye View (BEV); Bi-Level Routing Attention; Sparse Transformer
\end{IEEEkeywords}
\section{Introduction}
Compared with costly LiDAR-based solutions, vision based multi-view 3D object detection has garnered extensive attention due to its low hardware deployment cost and rich semantic information \cite{wang2021fcos3d, wang2022detr3d}. 
In particular, the Bird's Eye View (BEV) \cite{wang2025developments} perception enables the unified transformation of 2D image features from multiple on-board cameras into an ego-vehicle-centric 3D space \cite{huang2021bevdet}. With the comprehensive breakthrough of the Transformer architecture,  attention-based BEV perception algorithms have been proposed successively.
Despite the accuracy improvements achieved by existing Transformer-based multi-view perception algorithms, they still face severe challenges when deployed on practical on-board edge devices \cite{wang2025physically}. 
Firstly, the computational complexity of global attention generally grows quadratically, resulting in computational overhead. 
Secondly, existing multi-scale feature fusion modules often suffer from severe computational redundancy \cite{lin2017feature,zhu2020deformable}, and exhibit insufficient capability in preserving features of distant small-scale objects. 
More critically, mainstream 2D-to-3D view projection mechanisms universally adopt dense spatial grid queries and cross-attention computation. 
Such dense projection not only consumes massive computing resources, but also causes the model to perform ineffective interactions in meaningless background regions, which constrains the real-time inference performance \cite{li2024fast}.

To address the limitations, this paper proposes an efficient 3D object detection framework Sparse-BEVNet for autonomous driving. Specifically, Bi-Level Routing Attention (BRA) \cite{zhu2023biformer} is introduced into the image feature extraction stage. BRA dynamically filters irrelevant backgrounds at the coarse-grained region level to reduce the computational burden while enhancing the feature capture capability for small objects. In the multi-scale feature fusion stage, a feature neck based on Cascaded Group Attention (CGA) is constructed. CGA divides features of different scales into multiple groups for cascaded deep interaction to improve the feature consistency of multi-view overlapping areas and multi-scale objects. To break through the computational bottleneck of view transformation \cite{lin2026sparse4d}, this paper further proposes a Sparse Spatial Cross-Attention mechanism SSCA. It abandons traditional dense BEV grid queries and adopts instance-based sparse 3D reference points.
To verify the effectiveness of Sparse-BEVNet, experimental evaluations are conducted on the nuScenes public dataset \cite{caesar2020nuscenes}. Extensive quantitative comparisons show that the proposed method achieves a mean Average Precision (mAP) of 45.2\% and a nuScenes Detection Score (NDS) of 54.5\%, representing significant improvements of 3.6\% and 2.8\% respectively over the baseline algorithm.
\section{Methodology}

\subsection{Overall Network Architecture}

\begin{figure}[htbp]
	\centering
	\includegraphics[width=0.5\textwidth]{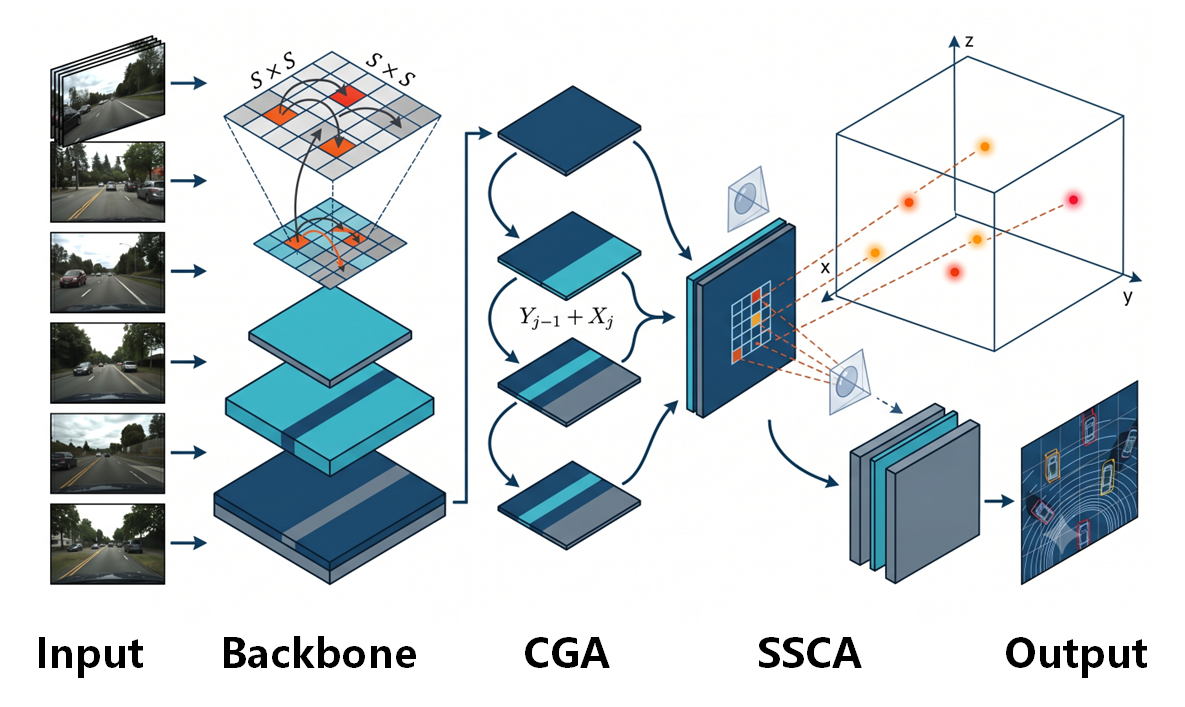}
	\caption{Overall architecture of the proposed Sparse-BEVNet.}
	\label{fig:overall_arch}
\end{figure}

The proposed Sparse-BEVNet is designed to address the problems of high computational cost and insufficient multi-scale feature capture in conventional multi-view single-frame perception algorithms. This network is s developed via modular reconstruction based on the well-established Sparse4D and BEVFormer \cite{li2022bevformer}. As illustrated in Fig.~\ref{fig:overall_arch}, the overall architecture consists of four modules: feature extraction backbone based on BiFormer (Backbone), multi-scale feature fusion neck based on cascaded group attention (CGA),  sparse spatial  cross-attention mechanism, and a detection head. 

\subsection{Backbone Feature Extraction Based on Bi-Level Routing}

Vanilla baseline methods generally adopt ResNet-101 or Swin Transformer as the feature extraction backbone. However, the local convolution operations of ResNet struggle to capture long-range dependencies, while the fixed-window attention mechanism of Swin not only fragments spatial context, but also assigns equal computational weights to large-area low-information-entropy backgrounds (e.g., sky and road surfaces) in high-resolution surround-view images, resulting in severe waste of computing resources.

To address this issue, this paper reconstructs the original backbone network into a lightweight feature extractor based on the BiFormer architecture, and introduces the Bi-Level Routing Attention (BRA) mechanism. 
Specifically, given a 2D feature map output from the shallow network $X \in \mathbb{R}^{H \times W \times C}$, the network first uniformly divides it into $S \times S$ non-overlapping macro regions along the spatial dimension. To evaluate the semantic correlations between different regions with extremely low computational overhead, the algorithm generates dimension-reduced region-level queries $Q^r$ and region-level keys $K^r \in \mathbb{R}^{S^2 \times C}$ by performing average pooling projection on tokens within each region.

Subsequently, the network computes the coarse-grained semantic affinity matrix between macro regions, and applies an asymmetric truncation operation to extract the top-$k$ most relevant region indices, thereby constructing a directed sparse region dependency graph:
\[
I^r = \operatorname{TopK}\left(Q^r (K^r)^\top, k\right)
\]
where $I^r \in \mathbb{N}^{S^2 \times k}$ denotes the generated routing index matrix.

Based on this dynamic routing decision, for the $i$-th specific region, the network no longer blindly interacts with pixels in the global scope or a fixed window. Instead, it only aggregates corresponding key and value tokens from the $k$ highly relevant regions pointed to by $I^r_i$ via the Gather operation, and reconstructs the dynamic context tensor:
\[
K^g_i = \operatorname{Gather}\left(K, I^r_i\right), \quad V^g_i = \operatorname{Gather}\left(V, I^r_i\right)
\]

After completing the above tensor reconstruction, fine-grained attention computation is only performed between query tokens of the current region and the dynamically collected context tensors $(K^g_i, V^g_i)$.

\subsection{Feature Fusion Based on Cascaded Group Attention}

After multi-level feature extraction, baseline models generally adopt the standard Feature Pyramid Network (FPN) for simple feature concatenation and fusion. This approach suffers from redundant attention computation and weak cross-scale interaction, making it difficult to cope with the drastic scale variation of objects at different distances in autonomous driving scenarios. To this end, this paper replaces the original FPN structure with a feature fusion neck based on Cascaded Group Attention (CGA), which aims to significantly enhance the deep interaction and representation capability of multi-scale features without introducing additional parameter overhead.

Specifically, the CGA module first uniformly divides the input multi-level feature maps into $M$ independent group subsets $X = [X_1, X_2, \dots, X_M]$ along the channel dimension, and allocates a separate attention head to each group for computation. To break the limitation of information isolation among groups, CGA introduces an explicit cascaded connection mechanism, which injects the attention output features of the preceding group as additional contextual information into the input of the current group. The mathematical formulation of this cascaded fusion process can be formalized as:
\[
Y_j = \operatorname{Attn}_j\left(X_j + Y_{j-1}\right), \quad j \in \{1, 2, \dots, M\}
\]
where $\operatorname{Attn}_j$ denotes the attention operation applied to the $j$-th group, $Y_j$ represents the output feature of the $j$-th group, and the initial boundary condition is set as $Y_0 = 0$.

Through this progressively accumulated serial computation path, each subsequent group in the network can not only focus on its own exclusive feature subspace, but also continuously absorb and integrate the receptive field information of all preceding groups. Finally, the output features $Y_j$ of all groups are re-concatenated along the channel dimension, and processed by linear projection to generate an enhanced high-quality multi-scale 2D feature set $F$.%

\subsection{Sparse Spatial Cross-Attention}

Existing BEV perception algorithms typically construct dense grid queries in 3D space to accomplish the 2D-to-3D view transformation. However, the vast majority of the 3D physical space in autonomous driving scenarios is vacant. Dense grid projection not only consumes massive computing resources, but also inevitably introduces a large amount of ineffective background noise interactions. To address this issue, this paper proposes a pure and efficient sparse spatial cross-attention mechanism SSCA, which replaces implicit dense grid search with deterministic geometric physical mapping.

Specifically, the network first initializes a set of sparse instance query features $Q$ and their corresponding 3D reference points $P = (x, y, z)$ in the unified 3D coordinate system. To establish precise associations between these 3D anchors and the previously generated multi-scale 2D image features $F$, the proposed method performs geometric projection using the actual physical calibration parameters of on-board cameras. 
For the $i$-th camera with intrinsic matrix $K_i \in \mathbb{R}^{3 \times 3}$ and extrinsic transformation matrix $T_i \in \mathbb{R}^{3 \times 4}$, the projection coordinate $p_i = (u, v)$ of a 3D reference point onto the corresponding 2D pixel plane can be rigorously computed via homogeneous coordinate mapping.

After obtaining the accurate 2D pixel coordinates, the network no longer performs time-consuming global feature matching. To adapt to object scale variations and projection errors, this paper formalizes the Sparse Spatial Cross-Attention (SSCA) as a process of local deformable sampling and feature aggregation based on projection coordinates:
\[
\operatorname{SSCA}(Q, P) = \sum_{i \in \mathcal{V}(P)} \sum_{l=1}^{L} \sum_{k=1}^{K} W_{i,l,k} \cdot F_{i,l}\left(p_i + \Delta p_{i,l,k}\right)
\]

In this formulation, $\mathcal{V}(P)$ represents the set of visible camera views where the 3D reference point $P$ can be effectively projected, i.e., without exceeding the image boundary or being occluded by the ego-vehicle. $L$ denotes the number of multi-scale feature levels, and $K$ is the number of local reference points sampled at each level. $\Delta p_{i,l,k}$ and $W_{i,l,k}$ are the 2D spatial sampling offsets and normalized attention weights respectively, both predicted from the query feature $Q$ via linear layers. $F_{i,l}$ refers to the 2D feature map of the corresponding camera and level output by the CGA neck.


After completing the 2D-to-3D feature lifting, the sparse query sequence enriched with abundant visual features is directly fed into the task detection head. The detection head consists of a set of parallel multi-layer perceptrons (MLPs), which decode and regress each sparse instance in parallel to output the final 3D bounding box parameter matrix. The state vector of each predicted object is expressed as $\hat{y} = (x, y, z, w, l, h, \theta, c)$, corresponding to the 3D center coordinates, physical dimensions (width, length, height), yaw angle, and confidence distribution of object categories in the ego-vehicle coordinate system, respectively. In this way, an end-to-end closed loop of single-frame 3D environmental perception is established.

\section{Experiments}
\subsection{Experimental Settings}

The experiments are conducted on nuScenes, a highly authoritative multi-view perception dataset in the autonomous driving field. This dataset covers diverse complex environments including sunny days, rainy days, and nighttime, and provides synchronized images from six surround-view cameras mounted around the ego-vehicle. Considering the computational constraints of experimental hardware and the efficiency of model iterative validation, the complete full dataset is not employed in this work. Instead, a highly representative subset is selected via uniform sampling for both training and evaluation. This subset consists of 200 real road scenes, which is divided into a training set and a validation set following the 8:2 ratio.

The development and testing of the proposed algorithm are implemented based on mmdetection3d, an open-source autonomous driving perception framework, and the PyTorch deep learning engine. All model training and inference evaluations are conducted on a high-performance server cluster equipped with 4 NVIDIA A100 Tensor Core GPUs (80 GB VRAM per GPU), which fully meets the parallel computing demands of multi-view high-resolution images.The experiments adopt mean Average Precision (mAP) and nuScenes Detection Score (NDS) as evaluation metrics.In terms of inference efficiency, Frames Per Second (FPS) is adopted as the sole metric to measure the real-time performance of the algorithm.

In the training phase, the input images from the 6 surround-view cameras are uniformly resized and cropped to a resolution of $704 \times 256$ pixels. The AdamW optimizer is adopted for model optimization, with an initial learning rate of $2 \times 10^{-4}$ and a cosine annealing strategy for dynamic learning rate decay. The weight decay coefficient is set to $1 \times 10^{-2}$. The entire network is trained for 100 epochs with a global batch size of 8.

\subsection{Comparative Analysis of Main Experiments}
To verify the performance of the proposed Sparse-BEVNet in single-frame 3D object detection, quantitative comparisons with mainstream multi-view detection algorithms are conducted on the nuScenes validation subset, with detailed results presented in Table \ref{tab:main_comparison}.

\begin{table}[htbp]
	\centering
	\caption{Performance comparison of different multi-view 3D object detection algorithms on the nuScenes validation set}
	\begin{tabular}{lcccc}
		\hline
		Method & Backbone  & mAP (\%) & NDS (\%) & FPS \\
		\hline
		FCOS3D & ResNet-101  & 29.5 & 37.2 & 12 \\
		DETR3D & ResNet-101  & 30.3 & 37.4 & 14 \\
		BEVFormer-S & ResNet-101  & 41.6 & 51.7 & 15 \\
		\textbf{Ours} & \textbf{BiFormer}  & \textbf{45.2} & \textbf{54.5} & \textbf{22} \\
		\hline
	\end{tabular}
	\label{tab:main_comparison}
\end{table}

As shown in Table \ref{tab:main_comparison}, Sparse-BEVNet achieves 45.2\% $mAP$ and 54.5\% $NDS$, surpassing the BEVFormer-S baseline by 3.6\% and 2.8\% absolute improvements respectively. The accuracy gain stems from two core designs: bi-level routing attention filters invalid backgrounds early in feature extraction to boost the signal-to-noise ratio of key instances, and cascaded group attention integrates cross-view truncated features to improve the recall of hard samples like distant pedestrians.

For inference efficiency, limited by the cross-attention computation of global dense grids, the conventional BEVFormer algorithm only achieves a single-frame inference speed of 15 FPS, which is difficult to cope with unexpected road conditions under high-speed driving. In contrast, the proposed Sparse-BEVNet completely abandons the dense 2D-to-3D view projection mechanism, and directly maps 3D sparse anchors to the 2D plane for local sampling via camera intrinsic and extrinsic matrices. This pure spatial sparse interaction paradigm based on geometric priors not only successfully avoids the massive computational waste caused by vast sky areas and static road surfaces, but also greatly boosts the inference frame rate of the model to 22 FPS. 

\subsection{Ablation Experiments and Analysis}

To quantify the individual contributions of the three core components ablation experiments are conducted on the nuScenes validation subset under consistent experimental settings. The baseline is built with ResNet-101 backbone, standard FPN neck and dense BEV grid projection.

\begin{table}[htbp]
	\centering
	\caption{Ablation results of core components in Sparse-BEVNet}
	\begin{tabular}{cccccc}
		\hline
		BRA & CGA & SSCA  & mAP (\%) & NDS (\%) & FPS \\
		\hline
		&  &  & 41.6 & 51.7 & 15 \\
		$\checkmark$ &  &  & 42.8 & 52.6 & 17 \\
		$\checkmark$ & $\checkmark$ &  & 43.7 & 53.4 & 16 \\
		$\checkmark$ & $\checkmark$ & $\checkmark$ & \textbf{45.2} & \textbf{54.5} & \textbf{22} \\
		\hline
	\end{tabular}
	\label{tab:ablation}
\end{table}

As shown in Table \ref{tab:ablation}, the three modules bring complementary performance gains when added sequentially. 
First, introducing BRA alone improves mAP by 1.2\% and raises inference speed to 17 FPS.
Specifically, compared to the heavy ResNet-101 baseline (comprising approx. 44.5M parameters and 7.8 GFLOPs), the proposed BRA-based backbone significantly curtails the computational overhead to roughly 26.0M parameters and 4.5 GFLOPs. 
Second, the cross-scale interaction of CGA compensates for feature truncation in multi-view overlapping areas and feature loss of distant small objects. Adding CGA further brings an extra 0.9\% mAP improvement.  Finally, replacing dense grid queries with SSCA achieves optimal accuracy and lifts inference speed to 22 FPS, demonstrating that geometry-guided sparse projection cuts redundant computation in empty regions.

Furthermore, to investigate the impact of different multi-scale merging paths in the feature fusion module, an additional ablation study was conducted on the input resolution tiers of the CGA neck. As shown in Table~\ref{tab:fusion_path}, fusing only the deep semantic layers ($\{C_4, C_5\}$) yields sub-optimal detection scores due to the severe loss of geometric details. While adding the intermediate layer ($\{C_3, C_4, C_5\}$) improves the performance, the proposed full hierarchical cascade ($\{C_2, C_3, C_4, C_5\}$) achieves the highest mAP and NDS. This confirms that integrating the high-resolution shallow features is crucial for detecting truncated and small targets, representing the optimal trade-off in the architectural design.

\begin{table}[htbp]
	\centering
	\caption{Comparison of alternative multi-scale merging paths in the CGA neck}
	\label{tab:fusion_path}
	\begin{tabular}{lcc}
		\hline
		Fusion Resolution Tiers & mAP (\%) & NDS (\%) \\
		\hline
		Deep layers only $\{C_4, C_5\}$ & 40.8 & 51.0 \\
		Partial cascade $\{C_3, C_4, C_5\}$ & 43.1 & 52.8 \\
		Full cascade $\{C_2, C_3, C_4, C_5\}$ (Proposed) & \textbf{45.2} & \textbf{54.5} \\
		\hline
	\end{tabular}
\end{table}
\section{Conclusion}
To address computational redundancy and insufficient multi-scale feature extraction in multi-view 3D object detection for autonomous driving, this paper proposes Sparse-BEVNet.
With bi-level routing attention (BRA) for backbone reconstruction and cascaded group attention (CGA) as the feature neck, it filters background noise effectively and enhances feature representation for cross-scale and truncated objects.
The sparse spatial cross-attention (SSCA) further breaks the computational bottleneck of conventional dense view projection in  2D-to-3D geometric mapping.
Experiments on the nuScenes dataset show that the proposed method outperforms the baseline in both mAP and NDS.

\bibliographystyle{IEEEtran}
\bibliography{references}

\end{document}